\documentclass[letterpaper]{article}
\usepackage{aaai18}
\usepackage{times}
\usepackage{url}            % simple URL typesetting
\usepackage{amsmath}
\usepackage{amssymb,amsmath,amsthm,enumitem}
\usepackage{helvet}
\usepackage{courier}
\usepackage{amssymb}
\usepackage{verbatim}
\usepackage{graphicx} % Allows including images
\usepackage[ruled,vlined]{algorithm2e}
\usepackage[normalem]{ulem}

\usepackage{xcolor}

\begin{document}
%\title{ Variational Learning of Global Latent Features for Text Sequence Matching}
%\title{ A Deconvolutional latent-variable model for Semi-supervised Sequence Matching}
\title{Deconvolutional Latent-Variable Model for Text Sequence Matching}
\author{Dinghan Shen, Yizhe Zhang, Ricardo Henao, Qinliang Su, Lawrence Carin \\
	Department of Electrical \& Computer Engineering,  Duke University \\
	Durham, NC 27708\\
	\{dinghan.shen, yizhe.zhang, ricardo.henao, qinliang.su, lcarin\}@duke.edu
}
\maketitle
 \begin{abstract}
	%\begin{quote}
%Determining semantic relationships between text sequences is an important step toward natural language understanding. 
%Fixed-length sentence embeddings are a scalable and efficient strategy, due to low matching complexity.
A latent-variable model is introduced for text matching, inferring sentence representations by jointly optimizing generative and discriminative objectives.
To alleviate typical optimization challenges in latent-variable models for text, we employ deconvolutional networks as the sequence decoder (generator), providing learned latent codes with more semantic information and better generalization.
Our model, trained in an unsupervised manner, yields stronger empirical predictive performance than a decoder based on Long Short-Term Memory (LSTM), with less parameters and considerably faster training.
Further, we apply it to text sequence-matching problems.
The proposed model significantly outperforms several strong sentence-encoding baselines, especially in the semi-supervised setting.
	%\end{quote}
\end{abstract}

\section{Introduction}
\smallskip
%\begin{comment}
%When analyzing the relationship between a pair of sentences, one often maps each into low-dimensional abstract representations (embeddings), and compares these mappings to infer relatedness or similarity. This task is important because it can be explicitly or implicitly generalized to many natural language processing (NLP) applications, such as paraphrase identification, textual entailment, information extraction, question answering, document summarization or machine translation (\cite{dagan2006pascal}). \par
%\end{comment}
The ability to infer the degree of match between two text sequences, and determine their semantic relationship, is of central importance in natural language understanding and reasoning \cite{bordes2014semantic}.
With recent advances in deep neural networks, considerable research has focused on developing \emph{end-to-end} deep learning models for text sequence matching \cite{hu2014convolutional,wang2016compare,rocktaschel2015reasoning,Wang:2017td,shen2017adaptive}.
State-of-the-art models typically first encode the text sequences into hidden units via a Long Short term Memory (LSTM) model or a Convolutional Neural Network (CNN), and techniques like attention mechanisms \cite{rocktaschel2015reasoning} or memory networks \cite{hill2015goldilocks} are subsequently applied for the final sequence matching, usually addressed as a classification problem.
However, the word-by-word matching nature of these models typically gives rise to high computational complexity, either $\mathcal{O}(T^2)$ \cite{wang2016compare} or $\mathcal{O}(T)$ \cite{rocktaschel2015reasoning}, where $T$ is the sentence length. Therefore, these approaches are computationally expensive and difficult to scale to large datasets or long text sequences. \par

Another class of models for matching natural language sentences is based on \emph{sentence encoding} methods, where each sentence is mapped to a vector (embedding), and two such vectors are used for predictions of relationships between the corresponding two sentences \cite{bowman2016fast,mou2015natural}.
In this case the matching complexity is independent of sentence length.
However, it has been found that is hard to encode the semantic information of an entire sequence into a single vector \cite{bowman2015large}.\par

For these models, it is important to learn an informative sentence representation with two properties:  (\emph{\romannumeral1}) it preserves its fundamental details, \emph{e.g.}, $n$-gram fragments within the sequence of text; (\emph{\romannumeral2}) the learned representation should contain discriminative information regarding its relationship with the target sequence.
So motivated, we propose to infer the embedding for each sentence with \emph{deep generative models}, due to their ability to make effective use of unlabeled data and learn abstract features from complex data \cite{kingma2014semi,yang2017improved,pu2016variational,wang2017zero}.
Moreover, the objective of a generative model addresses generation/reconstruction, and thus learns latent codes that naturally preserve essential information of a sequence, making them particularly well suited to sentence matching. \par

Recent advances in neural variational inference have manifested deep latent-variable models for text \cite{miao2016neural}.
The general idea is to map the sentence into a continuous latent variable, or \emph{code}, via an inference network (encoder), and then use the generative network (decoder) to reconstruct the input sentence conditioned on samples from the latent code (via its posterior distribution).
As a first attempt, \cite{bowman2016generating} proposed a Variational Auto-Encoder (VAE)-based generative model for text, with LSTM networks \cite{hochreiter1997long} as the sequence decoder.
However, due to the recurrent nature of the LSTM decoder, the model tends to largely ignore information from the latent variable; the learned sentence embedding contains little information from the input, even with several training modifications \cite{bowman2016generating}.
To mitigate this issue, \cite{yang2017improved} proposed to use a dilated CNN, rather than an LSTM, as a decoder in their latent-variable model.
Since this decoder is less dependent on the contextual information from previous words, the latent-variable representation tends to encode more information from the input sequence. \par

Unfortunately, regardless of whether LSTMs or dilated CNNs are used as the generative network, ground-truth words need to be fed into the decoder during training, which has two potential issues: (\emph{\romannumeral1}) given the powerful recursive and autoregressive nature of these decoders, the latent-variable model tends to ignore the latent vector altogether, thus reducing to a \emph{pure} language model (without external inputs)
\emph{i.e.}, latent representations are not effective during training \cite{bowman2016generating,chen2016variational}; (\emph{\romannumeral2}) the learned latent vector does not necessarily encode all the information needed to reconstruct the entire sequence, since additional guidance is provided while generating every word, \emph{i.e.}, \emph{exposure bias} \cite{ranzato2015sequence}. \par

We propose \emph{deconvolutional networks} as the sequence decoder in a latent-variable model, for matching natural language sentences.
Without any recurrent structure in the decoder, the typical optimization issues associated with training latent-variable models for text are mitigated.
Further, global sentence representations can be effectively learned, since no ground-truth words are made available to the decoder during training.
\par

In the experiments, we first evaluate our deconvolution-based model in an unsupervised manner, and examine whether the learned embedding can automatically distinguish different writing styles.
We demonstrate that the latent codes from our model are more informative than LSTM-based models, while achieving higher classification accuracy.
We then apply our latent-variable model to text-sequence matching tasks, where predictions are made only based on samples from the latent variables.
Consequently, without any prior knowledge on language structure, such as that used in traditional text analysis approaches (\emph{e.g.}, via a parse tree),  
%\rh{what do you mean?} \ds{[Modified a bit to make it more clear. I mean syntactic features that are widely adopted in traditional text analysis. This point is mentioned in \cite{hu2014convolutional}]},
our deconvolutional latent-variable model outperforms several competitive baselines, especially in the semi-supervised setting. \par

Our main contributions are as follows: \par
\emph{\romannumeral1}) We propose a neural variational inference framework for matching natural language sentences, which effectively leverages unlabeled data and achieves promising results with little supervision. \par 
\emph{\romannumeral2}) We employ deconvolutional networks as the sequence decoder, alleviating the optimization difficulties of training latent-variable models for text, resulting in more informative latent sentence representations. \par 
\emph{\romannumeral3}) The proposed deconvolutional latent-variable model is highly parallelizable, with less parameters and much faster training than LSTM-based alternatives.

\section{Background}
%
%\subsection{Encoding-based Sentence Matching}
\subsection{Matching natural language sentences}
Assume we have two sentences for which we wish to compute the degree of match.
For notational simplicity, we describe our model in the context of Recognizing Textual Entailment (RTE) \cite{rocktaschel2015reasoning}, thus we denote the two sequences as \emph{P} for premise and \emph{H} for hypothesis, where each sentence pair can be represented as $(p_i, h_i)$, for $ i = 1, 2, 3..., N$, where $N$ is the total number of pairs.
The goal of sequence matching is to predict judgement $y_i$ for the corresponding sentence pair, by modeling the conditional distribution $p(y_i|p_i, h_i)$, where $y_i\in\{$\emph{entailment}, \emph{contradiction}, \emph{neutral}$\}$.
\emph{Entailment} indicates that $p_i$ and $h_i$ can be inferred from each other, \emph{contradiction} suggests they have opposite semantic meanings, while \emph{neutral} means $p_i$ and $h_i$ are irrelevant to each other.
This framework can be generalized to other natural language processing applications, such as paraphrase identification, where $y_i=1$ if $p_i$ is a paraphrase of $h_i$, and $y_i=0$ otherwise. In this regard, text sequence matching can be viewed as either a binary or multi-class classification problem \cite{yu2014deep}. \par

Although word/phrase-level attention \cite{rocktaschel2015reasoning} or matching strategies \cite{wang2016compare} are often applied to text sequence-matching problems,  we only consider \emph{sentence encoding-based models}, because of their promising low complexity.
Specifically, our model is based on the \emph{siamese} architecture \cite{bromley1994signature},
%\sout{where two weight-sharing networks}
which consists of a twin network that 
%\rh{what are they sharing exactly, same architecture, same parameters, different data? if that is the case they really not sharing anything, the are the same network. Am I missing something here?}
processes natural language sentence pairs independently (the parameters of the twin network are tied); there is no interaction before both sentence representations are inferred.
%\ds{Yes, you are right. The two networks are exactly the same. So, I modified this part to use the description language from the original paper, what do you think?}
A classification layer is built on top of the two latent representations, for final prediction (matching).

The shared encoder network can be designed as any form of nonlinear transformation, including Convolutional Neural Networks (CNNs), Recurrent Neural Networks (RNNs) or Multi-Layer Perceptrons (MLPs).
However, to effectively match natural language sentences with the \emph{siamese} architecture, the key is to learn informative sentence representations through the encoder network.
%to learn informative and abstract sentence representations through the encoder network is %crucial to an effective siamese model for matching natural language sentences.
%\rh{incomplete idea, reads disconnected from the rest of the section.} \ds{modified. The logic is here is that good sentence representations are vital for the success of siamese networks, and in the next part I 'll talk about how to achieve this goal with latent-variable models. }
To this end, below we describe use of CNNs in the context of a latent-variable model. \par

\subsection{Latent-variable models for text processing}
% why we would want a vae model rather than a seq2seq model %
Sequence-to-sequence models \cite{sutskever2014sequence} are the most common strategy for obtaining robust sentence representations, as these are capable of leveraging  information from unlabeled data. These models first encode the input sentence $x$ (composed of $T$ words, $w_{1:T}$) into a fixed-length vector $z=g(x)$, and then reconstruct/generate the output sequence from $z$.
Specifically, in the autoencoder setup, the output of the decoder is the reconstruction of the input sentence $x$, denoted $\hat{x}$ with words $\hat{w}_{1:T}$,
\begin{align}
	p(\hat{x}|x) = & \ p(\hat{w}_{1:T}|w_{1:N}) \label{eq:seq2seq} \\
	= & \ p(\hat{w}_1|z=g(x)) \prod_{t=2}^{T} p(\hat{w}_t|z = g(x), \hat{w}_{1:t-1}) \,, \nonumber
\end{align}
%
%\rh{this equation has problems when $t=1$, it probably starts with $t=2$ and has a $p(\hat{x}_t|h)$ term} \ds{you are totally correct, I have modified it. Actually, I use the expression from equation (1) in this highly cited paper: https://papers.nips.cc/paper/5346-sequence-to-sequence-learning-with-neural-networks.pdf, however it turns to be wrong. }.
where $g(\cdot)$ is a {\em deterministic}, generally nonlinear transformation of $x$. The deterministic $g(x)$ may result in poor model generalization, especially when only a limited number of labeled data are available for training.
Below we consider a {\em probabilistic} representation for $z$, \emph{i.e.,} $p(z|x)$.
%\rh{I changed the notation above from $h$ to $z$, to match below. No need to have different notations for deterministic and stochastic} \ds{Agree. I think it looks good now.}

Recently \cite{miao2016neural} introduced a Neural Variational Inference (NVI) framework for text modeling, in which they infer a stochastic latent variable $z \sim q(z|x)$ to model the input text, constructing an inference network to approximate the true posterior distribution $p(z|x)$.
%\rh{how is $z$ learning the semantics? also you changed the notation, what happened with $h$? and the subscripts of $x$?}. \sout{In others words, the posterior distribution of sentence representaion $h$ is modeled} \rh{are you sure? do you mean the posterior of $z$?} \ds{modified}, \sout{endowing the latent representations with better generalization property}
This strategy endows latent variable $z$ with a better ability to generalize \cite{miao2016neural}.
Conditioning on the latent code $z$, a decoder network $p(x|z)$ maps $z$ back to reconstruct the original sequence, $x$.
Given a set of observed sentences (training set), the parameters of this model are learned by maximizing the marginal $p(x)$.
Since this is intractable in most cases, a variational lower bound is typically employed as the objective to be maximized \cite{kingma2013auto}:
%\rh{Note that you suddenly went from $x_{1:N}$ to $x$. Consider saying that a sentence is denoted as $x$ and it is composed of $T$ words, $w_{1:T}$ above.} \ds{modified exactly according to your suggestion.}
%
\begin{align}
  \mathcal{L}_{\rm vae} = & \ \mathbb{E}_{q_\phi(z|x)}[\log p_\theta(x|z)] - D_{KL}(q_\phi(z|x)|p(z)) \nonumber \\
  %\end{aligned} \notag \\
  %\begin{aligned}
   = & \ E_{q_\phi(z|x)} [\log p_\theta(x|z) + \log p(z)- \log q_\phi(z|x)] \nonumber \\
  %  \end{aligned} \notag \\
  \leq & \ \log \int p_\theta(x|z)p(z)dz = \log p_\theta(x) \,,
  \label{eq:vae}
\end{align}
%
%\rh{are you sure that $p_\theta(z)$ is parameterized by $\theta$? I don't think so.}
where $\theta$ and $\phi$ denote decoder and encoder parameters, respectively.
%Typically, $q_\phi(z|x)$ assumes a parametric form, \emph{e.g.}, a diagonal Gaussian and the re-parametrization trick \cite{kingma2013auto} is applied to facilitate training.
The lower bound $\mathcal{L}_{\rm vae}(\theta,\phi; x)$ is maximized w.r.t. both encoder and decoder parameters.
Intuitively, the model aims to minimize the reconstruction error as well as to regularize the posterior distribution $q_\phi(z|x)$ as to not diverge too much from the prior $p(z)$.
This neural variational inference framework has achieved significant success on other types of data, such as images \cite{gregor2015draw,pu2016variational}.
% \ds{maybe I could separate the section below to discuss the difficulty of training latent-variable models for text, to highlight the motivation for using deconvolution networks, what do yo think?} \rh{I like it, please add a small paragraph after the first sentence below that described why NVI has not be used as much for text by highlighting the difficulty of it.} \par

\subsection{Challenges with the NVI framework for text}
Extracting sentence features for text with the above NVI framework has been shown to be difficult \cite{bowman2016generating,yang2017improved}. For an unsupervised latent-variable model, which is often referred to as a variational autoencoder \cite{kingma2013auto}, the parameters are optimized by minimizing the reconstruction error of sentences, as well as regularizing the posterior distribution $q_\phi(z|x)$ to be close to the prior $p(z)$, as in \eqref{eq:vae} via $D_{KL}(q_\phi(z|x)|p(z))$.
Therefore, we can think of the variational autoencoder as a regularized version of a standard (deterministic) autoencoder (sequence-to-sequence model), due to the additional penalty term coming from KL divergence loss.\par

Although the KL divergence in \eqref{eq:vae} term plays a key role in training latent-variable models with the NVI framework, it has been reported that, when applied to text data (sentences), the KL loss tends to be insignificantly small during training \cite{bowman2016generating}.
As a result, the encoder matches the Gaussian prior regardless of the input, and the decoder doesn't take advantage of information from the latent variable $z$.
%Thus, the model reduces to be a language model.
Moreover, it has been reported that poor results in this setting may be attributed to the autoregressive nature of the LSTM decoder \cite{chen2016variational,bowman2016generating}.
While decoding, the LSTM imposes strong conditional dependencies between consecutive words, thus, from \eqref{eq:seq2seq}, the information from $z$ becomes less impactful during learning.
%\rh{what do you mean?} \ds{the same as before, LSTM has very strong conditional dependency at every times step, and thus ignore the information from latent variables}
%\rh{you mean a pure language model? note that before you said it was not a language model for the exact same reason.} \ds{as noted above, I misunderstood the meaning of 'cease to be', sorry about this mistake. }
Motivated by these issues, \cite{yang2017improved} employed dilated CNNs, instead of the LSTM, as a sentence decoder for a latent-variable model.
In \cite{yang2017improved} the latent variable $z$ is able to encode more semantic information, because of the smaller contextual capacity of the dilated CNN decoder.
%In particular, \cite{yang2017improved} use the the proportion of the loss due to the KL %term in the variational objective as a way to quantify the amount of information encoded %in the latent variable, $z$.
% which can be quantified by a larger proportion of KL divergence in the total loss \rh{not sure what you mean here} \ds{since the model tends to ignore latent variable $z$, the KL loss is typically very close to 0, which means no information is encoded in $z$. Therefore, the two papers cited above use the proportion of KL loss as a metric for how much information is encoded in $z$}, 
However, optimization challenges remain, because ground-truth words are employed while training, 
as the dilated CNN is an autoregressive decoder.
Consequently, the inferred latent codes cannot be considered as global features of a sentence, since they do not necessarily encode all the information needed to reconstruct an entire sequence.

\section{Model}
\subsection{Deconvolutional sequence decoder}
Deconvolutional networks, also known as \emph{transposed} convolutional layers, are typically used in deep learning models to up-sample fixed-length latent representations or high-level feature maps \cite{zeiler2010deconvolutional}.
Although widely adopted in image generative models, deconvolutional networks have been rarely applied to generative models for text. To understand the form of the decoder needed for text, we first consider the associated convolutional encoder (\cite{kim2014convolutional}, \cite{zhang2017deconvolutional}). The text is represented as a matrix, with ``width'' dictated by the sentence length and ``height'' dictated by the dimensionality of the word embeddings. With $K_1$ convolutional filters at layer 1 of the model, after one-dimensional (1D) convolution between the 2D filters and 2D sentence embedding matrix (convolution in the direction of the word index, or ``time''), $K_1$ 1D signals are manifested. Using these $K_1$ 1D feature maps, a similar process repeats to substantiate subsequent layers in the deep model. Hence, at layer $l$ of the model, there are $K_l$ 1D signals manifested from $K_l$ 1D convolutions between $K_l$ 2D filters and the 2D feature-map from layer $l-1$. 
%The decoding process seeks to use the latent code $z$ at the top of the model, and to sequentially recover $K_l$ 1D signals at each layer $l$, and at the bottom of the network we seek to estimate the sentence-embedding matrix.
\par

\begin{figure}
	\centering
	\includegraphics[scale=0.3]{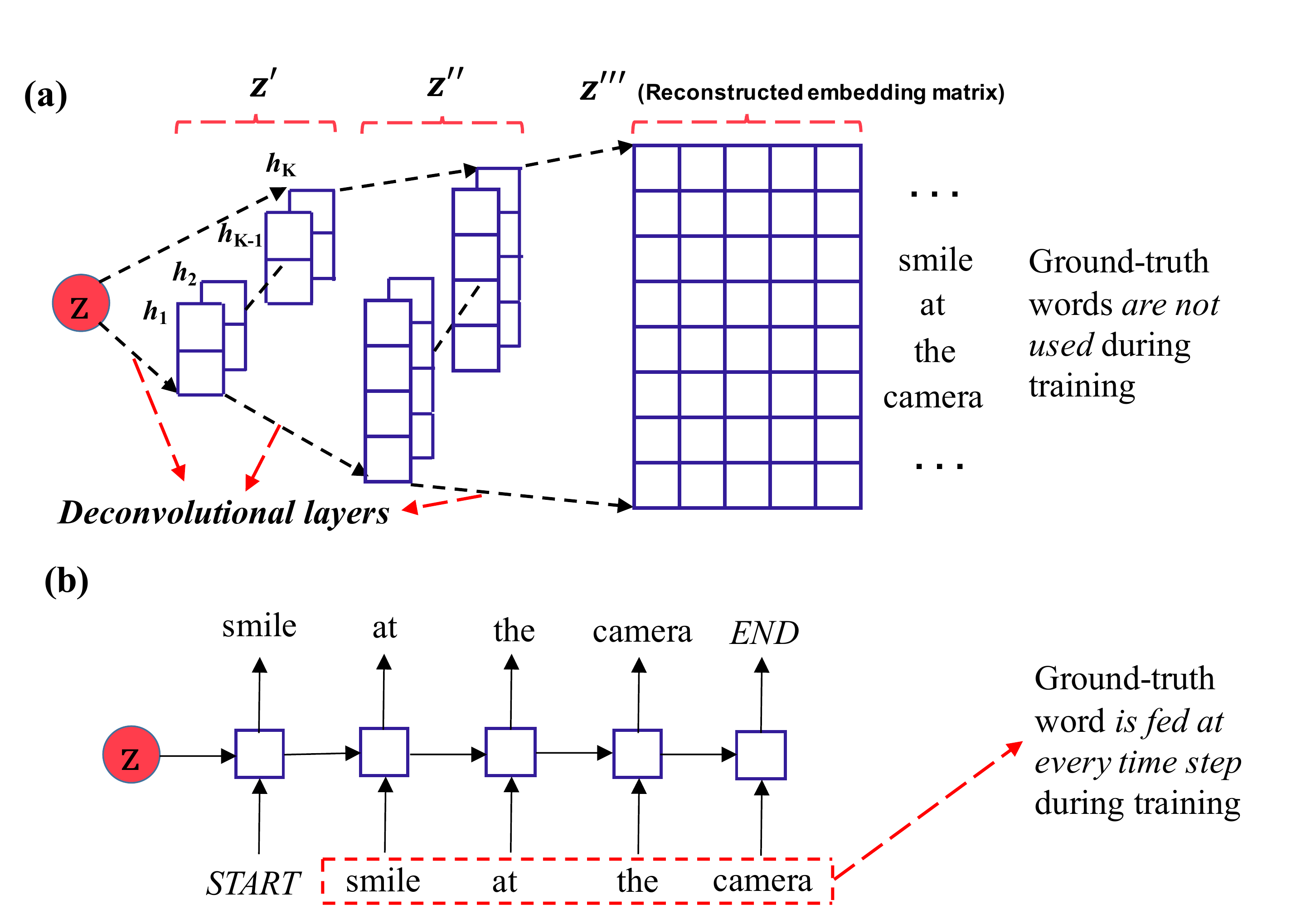}
	%\vspace{-5mm}
	\caption{\textbf{(a)} Diagram of deconvolutional sequence decoder, comparing with \textbf{(b)} LSTM sequence decoder. Notably, in contrast to a LSTM decoder, ground truth words are not provided for the deconvolutional networks during training. As a result, the failure mode of optimization described in \cite{bowman2016generating}, where the KL divergence term is vanishingly small, is largely mitigated. }
	\label{fig:deconv}
		%\vspace{-5mm}
\end{figure}
The encoder discussed above starts at the ``bottom'' with the sentence-embedding matrix, and works upward to the latent code $z$. The decoder works downward, starting at $z$ and arriving at the sentence-embedding matrix.
%Consider a one-layer deconvolutional network, employed as the sentence decoder. %The single-layer model is convenient for exposition, and in experiments we also consider a three-layer convolutional network as the sentence encoder. 
Specifically, the decoder network takes as input $z\in\mathbb{R}^M$ sampled from the inference (encoder) network $q_\phi(z|x)$.  
% \textcolor{blue}{Explicitly state how $q_\phi(z|x)$ is implemented in practice, to make explicit and clear.} \ds{addressed}
For an $L$-layer decoder model, the feature maps at layer $L$ (just beneath the latent code $z$) are manifested by $K_L$ filter matrices
% \textcolor{blue}{which network?} \ds{addressed}
$f_i^{(L)}\in\mathbb{R}^{H_L \times M}$, for $i = 1,2,....,K_L$, where $H_L$ corresponds to the number of components in the temporal (word) dimension. Each 2D matrix $f_i^{(L)}$ is multiplied by column vector $z$ (transpose convolution), yielding $K_L$ 1D feature maps. This yields an $H_L\times K_L$ feature-map matrix at layer $L$ (followed by ReLU pointwise nonlinearity). To yield the layer $L-1$ feature map matrix, the process repeats, using filters $f_i^{(L-1)}\in\mathbb{R}^{H_{L-1} \times K_L}$, for $i = 1,2,....,K_{L-1}$, with which $K_{L-1}$ 1D convolutions are performed with the feature-map matrix from layer $L$ (convolutions in the temporal/word dimension). This again yields a feature-map matrix at layer $L-1$, followed by ReLU nonlinearity. \par 

This process continues sequentially, until we arrive at the bottom of the decoder network, yielding a final matrix from which the sentence-embedding matrix is approximated. To be explicit, in Fig. \ref{fig:deconv} let $z^\prime$ and $z^{\prime\prime}$ represent the feature-map matrices at the top-two layers of a three-layer model. Let $z^{\prime\prime\prime}$ represent the matrix recovered at the bottom layer of the network through the above process, with ``height'' corresponding to the dimension of the word-embedding.
%%\rh{Overloaded, $N$ was used before to denote the number of sentence pairs.} \ds{replaced $N$ with $K$ to avoid notation overloading.}
%For the $i$-th filter $f_i$, we convolve it with the input representation $z$ to get an intermediate feature map, denoted as $h_i\in\mathbb{R}^{W \times H}$.
%Subsequently, all filter outputs are concatenated together to form the up-sampled representation, denoted here as $z^{\prime}\in \mathbb{R}^{W \times H \times K}$:
%%
%\begin{align}\label{eq:deconv}
%	%\begin{aligned}
%	h_i = \ z \circledast  f_{i}, \qquad
%	%\,\,\,\,\, i = 1,2,..., K \\
%%\end{gather}
%%\begin{gather}
%	z^{\prime} = \ h_1 \oplus \cdots \oplus h_i \oplus \cdots \oplus h_K \,,
%	%\end{aligned}
%\end{align}
%%
%where $\circledast$ is the convolutional operator
%%\rh{is this standard?} \ds{yep, either $\circledast$ or $\otimes$ can be used to denote convolution}
%and $\oplus$ denotes the concatenation operator. The general architecture of the deconvolutional network is shown in Figure~\ref{fig:deconv}(a). Further, for a two-layer decoder, $z^{\prime}$ is fed through \eqref{eq:deconv}, to obtain $z^{\prime\prime} \in \mathbb{R}^{W \times H \times K^{\prime}}$, then $z^{\prime\prime}$ is utilized to reconstruct the embedding matrix for sentence $x$. 
Suppose $\boldsymbol{E}$ is the word-embedding matrix for our vocabulary, and $\hat{w}_i$ the $i$th word in the reconstructed sentence. We compute the probability that $\hat{w}_i$ is word $s$ as:
\begin{align}\label{eq:cosine}
p(\hat{w}_i = s) = \frac{\textbf{\normalfont{exp}}\{\tau^{-1}\cos(z_{i}^{\prime\prime\prime}, 
	\boldsymbol{E}[s])\}}{\sum_{s^\prime \in V} \textbf{\normalfont{exp}}\{\tau^{-1}\cos(z_{i}^{\prime\prime\prime}, \boldsymbol{E}[s^\prime])\}} \,,
\end{align}
where $\cos(a, b)$ is the $\emph{\text{cosine similarity}}$ between vectors $a$ and $b$, $V$ is the vocabulary which contains all possible words and $\boldsymbol{E}[s]$ represents the column of $\boldsymbol{E}$ corresponding to word $s$; $z_{i}^{\prime\prime\prime}$ is the $i$-th column of the up-sampled representation $z^{\prime\prime\prime}$.
Parameter $\tau$ controls the sparsity of resulting probabilities, which we denote as the \emph{temperature} parameter.
We set $\tau=0.01$ in our experiments.
\par 
\par

The multilayer coarse-to-fine process (latent variable vector to embedding matrix) implied by repeatedly applying the above decoder process illustrated in Figure~\ref{fig:deconv}(a)) has two advantages:
%By replacing the sampled input latent variable with the previous layer's output, the deconvolutional layer described above can be hierarchically stacked to form a multilayer coarse-to-fine up-sampling process (as illustrated in Figure~\ref{fig:deconv}(a)).
%This hierarchical architecture has two advantages:
$i$) it reflects the natural hierarchical tree structure of sentences, thus may better represent syntactic features, which is useful when reconstructing sentences; $ii$) the deconvolutional network allows for efficient parallelization while generating each fragment of a sentence, and thus can be considerably faster than an LSTM decoder. 

%\subsubsection{Comparison with LSTM decoder}
As shown in  Figure~\ref{fig:deconv}, the training procedures for deconvolutional (a) and LSTM (b) decoders are intrinsically different. In the latter, ground-truth words of the previous time steps are provided while training the network. In contrast, the deconvolutional network generates the entire sentence (in block) from $z$ alone.
% \textcolor{blue}{How????}. \ds{I think the added part already addressed this issue}
%, as a sequence decoder, do not utilize any ground-truth words as input during training. 
Because of this distinction, the LSTM decoder, as an autoregressive model with powerful recurrence, tends to explain all structure in the data, with little insight from the latent variables which only provide information at the beginning of the sentence, thus acting merely as a prior.
%of the sentence (\cite{bowman2016generating}, \cite{chen2016variational}).
%Nevertheless, the deconvolutional neural networks, while generating a sentence, do not explicitly model the sequential dependency. In this regard, it could be particularly well suited to be employed, as the sequence decoder, to train latent-variable models for text data.

\subsection{Deconvolutional latent-variable models}
In this section we incorporate the deconvolutional sequence decoder described in the previous section in our latent-variable model for text.
Because of the coarse-to-fine generation process described above, the model does not have partial access to observed data (ground-truth words) during the generation process, as in an LSTM, thus the latent-variable model must learn to encode as much information as possible from the input alone.
% even though it may lead to larger KL loss penalties \cite{bowman2016generating}.
Moreover, in this way the learned latent code can be truly viewed as a global feature representation of sentences, since it contains all the essential information to generate the text sequence.
In the following, we describe the proposed deconvolutional latent-variable models, in the context of both unsupervised and supervised (including semi-supervised) learning.
% To better understand the proposed deconvolutional latent-variable model, we investigate it in the cases of both unsupervised and supervised (semi-supervised) learning. \rh{how is this helping us to understand the model?} \ds{I add this paragraph here to emphasize the motivation to utilize deconvolutional decoder for the latent-variable model. Maybe I should make it more clear and concisely describe it in a few sentences. Below is my modified version of this paragraph:}
% Motivated by the difficulties of training latent-variable models for text data, we propose to adopt deconvolutional networks as the sequence decoder. This model reconstructs the input sequence from latent variable $z$ in a coarse-to-fine manner, which allows $z$ to maintain as most information of the input as possible. In this section, we describe our deconvolutional latent-variable models, in the context of both unsupervised and supervised (semi-supervised) learning. \ds{what do you think?}

\begin{comment}
We first investigate the effectiveness of our model in the case of unsupervised learning, then we apply it to text sequence matching tasks.
\end{comment}

\subsubsection{Unsupervised sequence learning}
To demonstrate the effectiveness of our proposed model, we explore training it in an unsupervised manner. 
Specifically, for a input sentence $x$, the latent code is inferred through an encoder network $q_\phi(z|x)$ implemented as
\begin{align}
\mu = & \ g_1(f^{\rm cnn}(x;\phi_{10});\phi_{11}), & \log \sigma = & \ g_2(f^{\rm cnn}(x;\phi_{20});\phi_{21}) \nonumber \\
%\end{gather}
%\begin{gather}
\varepsilon \sim & \ \mathcal{N}(0, \bold{I}), &  z = & \ \mu + \varepsilon \odot  \sigma \,, \label{eq:define_latent_var}
\end{align}
where $f^{\rm cnn}(x;\phi_{10})$ denotes the transformation function of the encoder, accomplished via learning a CNN with input $x$ and parameters $\phi_{10}$, and $\odot$ represents the Hadamard vector product.
The posterior mean $\mu$ and variance $\sigma$ are generated through two non-linear transformations $g_1(\cdot)$ and $g_2(\cdot)$, both parameterized as neural networks; $g_1(y;\phi_{11})$ has input $y$ and parameters $\phi_{11}$.
Note that \eqref{eq:define_latent_var} is $q_\phi(z|x)$ in \eqref{eq:vae}, where $\phi=\{\phi_{10},\phi_{11},\phi_{20},\phi_{21}\}$.
Then $z$ is sampled with the re-parameterization trick \cite{kingma2013auto} to facilitate model training.
% \textcolor{blue}{[ref]}.  \ds{added} 
The sampled $z$ is then fed into a deconvolutional sequence decoder described above, to reconstruct the corresponding input sentences.
The model is trained by optimizing the variational lower bound in \eqref{eq:vae}, without any discriminative information.

\subsubsection{Supervised sequence matching}
\begin{figure}
	\centering
	%\vspace{0mm}
	\includegraphics[scale=0.45]{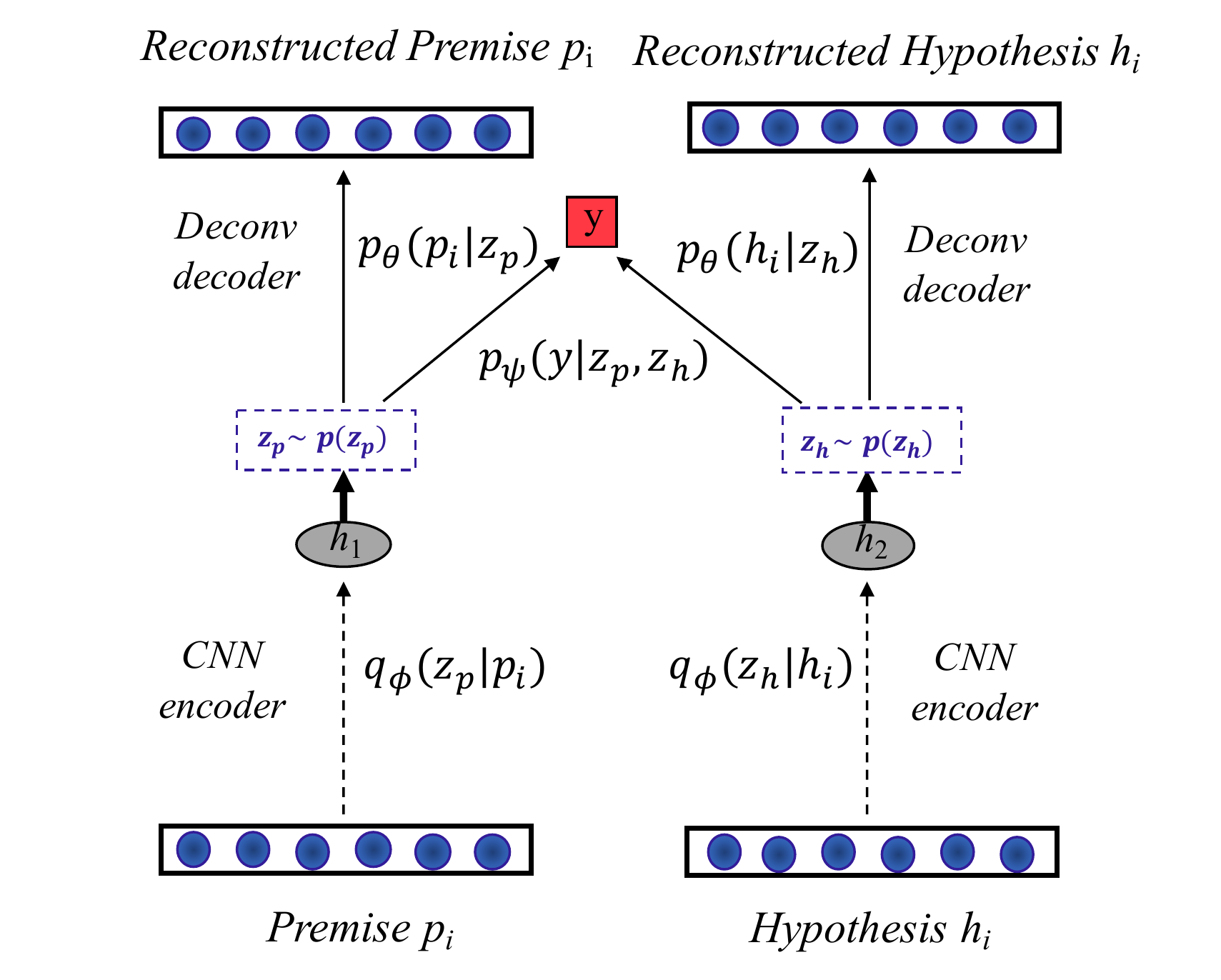}
	%\vspace{-2mm}
	\caption{Our deconvolutional latent-variable model for text sequence matching. The reconstruction/generation and discriminative objectives are jointly optimized to learn more robust latent codes for sentences.}
	\label{fig:vae}
	%\vspace{-4mm}
\end{figure}

We apply our latent-variable model to text sequence-matching problems, employing the discriminative information encoded in latent code $z$ (see Figure~\ref{fig:vae}).
For a sentence pair $(p_i, h_i)$, the latent code for each sequence is inferred as in \eqref{eq:define_latent_var}, where the parameters of the encoder network for $z_p$ and $z_h$, premise and hypothesis, respectively, are shared.
They are decoded by two shared-weight deconvolution networks, to recover the corresponding input sentence. \par

To infer the label, $y$, the two latent features are again sampled from the inference network and processed by a matching layer, to combine the information in the two sentences.
This matching layer, defined as \emph{heuristic} matching layer by \cite{mou2015natural}, can be specified as:
% Following \cite{mou2015natural}, the \emph{heuristic}
%\rh{why heuristic?} \ds{this matching layer is called heuristic matching layer in  \cite{mou2015natural}, that is why I use 'heuristic' here. One better word may be 'multi-perspective' matching layer, what do you think?}
% matching layer is defined as:
%  
\begin{align*}
m = [z_p; z_h; z_p - z_h; z_p \odot z_h] \,,
\end{align*}
%
% \sout{The four} \rh{which 4?)} \ds{I mean $z_1, z_2, z_1 - z_2, z_1 \circ z_2$, maybe I can rephrase it to ''the matching features $m$ is further fed to a classifier}
These matching features are stacked together into $m\in\mathbb{R}^{4M}$, for $z_p,z_h\in\mathbb{R}^{M}$, and fed into a classifier. The classifier is a two-layer MLP followed by a fully-connected softmax layer, that outputs the probabilities for each label (entailment, contradiction and neutral), to model the conditional distribution $p_\psi(y|z_p, z_h)$, with parameters $\psi$. \par

To allow the model to explore and balance between maximizing the variational lower bound and minimizing the sequence matching loss, a joint training objective is employed:
\begin{align}
%\mathcal{L}^{label} = -\sum_{i = 1, 2} {\mathcal{L}_{vae}(\theta,\phi; x_i)} + \alpha
\mathcal{L}^{\rm label} = - & \ {\mathcal{L}_{\rm vae}(\theta,\phi; p_i)} -  {\mathcal{L}_{\rm vae}(\theta,\phi; h_i)} \nonumber \\
 + & \ \alpha \mathcal{L}_{\rm match}(\psi; z_p, z_h, y) \,, \nonumber %\label{eq:label}
\end{align}
where $\psi$ refers to parameters of the MLP classifier
%\rh{please define the classifier} \ds{I think it has been defined as a two-layer MLP above}
and $\alpha$ controls the relative weight between the generative loss, $\mathcal{L}_{\rm vae}(\cdot)$, and sequence matching loss, $\mathcal{L}_{\rm match}(\cdot)$, defined as the cross-entropy loss.
When implementing this model, we anneal the value of $\alpha$ during training from 0 to 1 (the annealing rate is treated as a hyperparameter), 
%\rh{how, from where to where?} \ds{modified}
so that the latent variable learned can gradually focus less on the reconstruction objective, only retaining those features that are useful for sequence matching, \emph{i.e.}, minimizing the second term.

\subsubsection{Extension to semi-supervised learning}
Our latent-variable model can be readily extended to a semi-supervised scenario, where only a subset of sequence pairs have corresponding class labels.
Suppose the empirical distributions for the labeled and unlabeled data are referred to as $\tilde{p}_l(P, H, y)$ and $\tilde{p}_u(P, H)$, respectively.
The loss function for unlabeled data can be expressed as:  
\begin{align*}
%\mathcal{L}^{unlabel} = - \sum_{i = 1, 2} {\mathcal{L}_{vae}(\theta,\phi; x_i)}
\mathcal{L}^{\rm unlabel} = - {\mathcal{L}_{\rm vae}(\theta,\phi; p_i)}  - {\mathcal{L}_{\rm vae}(\theta,\phi; h_i)} \,.
\end{align*}
Therefore, the overall objective for the joint latent-variable model is:
%
%\begin{gather}
\begin{align}\label{eq:l_joint}
	\begin{aligned}
\mathcal{L}_{\rm joint} = & \ \mathbb{E}_{(p_i, h_i, y)\sim \tilde{p}_l}[\mathcal{L}^{\rm label}(p_i, h_i, y)] \\
%\end{aligned}  \notag\\
+ & \ \mathbb{E}_{(p_i, h_i) \sim \ \tilde{p}_u}[\mathcal{L}^{\rm unlabel}(p_i, h_i)] \,.
	\end{aligned}
\end{align}
To minimize $\mathcal{L}_{\rm joint}$ w.r.t. $\theta$, $\phi$ and $\psi$, we employ Monte Carlo integration to approximate the expectations in \eqref{eq:l_joint}. In this case unlabeled data are leveraged in the objective via the standard VAE lower bound. During training, all parameters are jointly updated with stochastic gradient descent (SGD).  
%\rh{so how do you optimize $\mathcal{L}_{\rm joint}$?}
% unlabeled data are made use of by maximizing their variational lower bound \rh{what do you mean?} \ds{I mean in our model, the unlabeled data is leveraged through directly optimizing the standard VAE lower bound.}

\section{Experiments}
\subsection{Experimental Setup}
Our deconvolutional latent-variable model can be trained in an unsupervised, supervised or semi-supervised manner.
In this section we first train the model in an unsupervised way, with a mixed corpus of scientific and informal writing styles, and evaluate the sentence embeddings by checking whether they can automatically distinguish different sentence characteristics, \emph{i.e.}, writing styles.
Further, we apply our models to two standard text sequence matching tasks: Recognizing Textual Entailment (RTE) and paraphrase identification, in a semi-supervised setting.
The summary statistics of both datasets are presented in Table~\ref{tab:summary}.

For simplicity, we denote our deconvolutional latent-variable model as DeConv-LVM in all experiments.
To facilitate comparison with prior work, several baseline models are implemented: (\emph{\romannumeral1}) a basic Siamese model with CNNs as the encoder for both sentences, with sharing configurations and weights; (\emph{\romannumeral2}) an auto-encoder with CNN as the sequence encoder and DeConv as decoder; 3) a latent-variable model using a CNN as the inference network, and the generative network is implemented as an LSTM (denoted LSTM-LVM).

%[network architecture]
We use 3-layer convolutional neural networks for the inference/encoder network, in order to extract hierarchical representation of sentences (\cite{hu2014convolutional}).
Specifically, for all layers we set the filter window size ($W$) as 5, with a stride of 2.
The feature maps ($K$) are set as 300, 600, 500, for layers 1 through 3, respectively.
In our latent-variable models, the 500-dimension feature vector is then fed into two MLPs to infer the mean and variance of the latent variable $z$.
The generative/decoder network is implemented as 3-layer deconvolutional networks, to decode the samples from latent variable $z$ of size $M=500$.
% We set the dimension of $z$ ($M$) to 500, the same as that of feature vector $h$.

\begin{table}
	\centering
	\begin{tabular}{cccccc}
		\hline
		Dataset & Train &Test & Classes &  Vocabulary\\
		\hline
		Quora       & 384348 & 10000 & 2 & 10k \\
		SNLI         & 549367 & 9824 & 3 &  20k \\
		\hline
	\end{tabular}
	\caption{Summary of text sequence matching datasets.}
	\label{tab:summary}
\end{table}
 
 %[training details]
The model is trained using Adam \cite{kingma2014adam} with a learning rate of $3 \times 10^{-4}$ for all parameters. 
Dropout \cite{srivastava2014dropout} is employed on both word embedding and latent variable layers, with rates selected from \{0.3, 0.5, 0.8\} on the validation set.
We set the mini-batch size to 32.
In semi-supervised sequence matching experiments, $L_2$ norm of the weight vectors is employed as a regularization term in the loss function, and the coefficient of the $L_2$ loss is treated as a hyperparameter and tuned on the validation set.
All experiments are implemented in Tensorflow \cite{abadi2016tensorflow}, using one NVIDIA GeForce GTX TITAN X GPU with 12GB memory.

\subsection{Unsupervised Sentence Embedding}
To investigate the effectiveness of our latent-variable model, we first train it in an unsupervised manner, using the dataset in \cite{zhang2017adversarial}, where sentences from two  corpora, \emph{i.e}, \emph{BookCorpus} dataset \cite{zhu2015aligning} and the \emph{arXiv} dataset, are merged together in equal proportion.
The motivation here is to check whether the latent codes learned in our model can automatically distinguish between different writing styles, \emph{i.e.}, sentences with scientific or informal styles represented by \emph{BookCorpus} and \emph{arXiv} dataset, respectively.
In this experiment, our model is trained by optimizing the variational lower bound in \eqref{eq:vae}, without any label/discriminative information provided.
We compare our model with another latent-variable model using LSTM as the decoder, to especially highlight the contribution of the deconvolutional network to the overall setup.
To ensure a fair comparison, we employ the same model architecture for the LSTM-based latent-variable model (LSTM-LVM), except for the decoder utilized.
The LSTM hidden-state dimension is set to 500, with the latent variable $z$ fed to decoder as input at every time step.
	
\begin{figure}
	\centering
	\includegraphics[width=.225\textwidth]{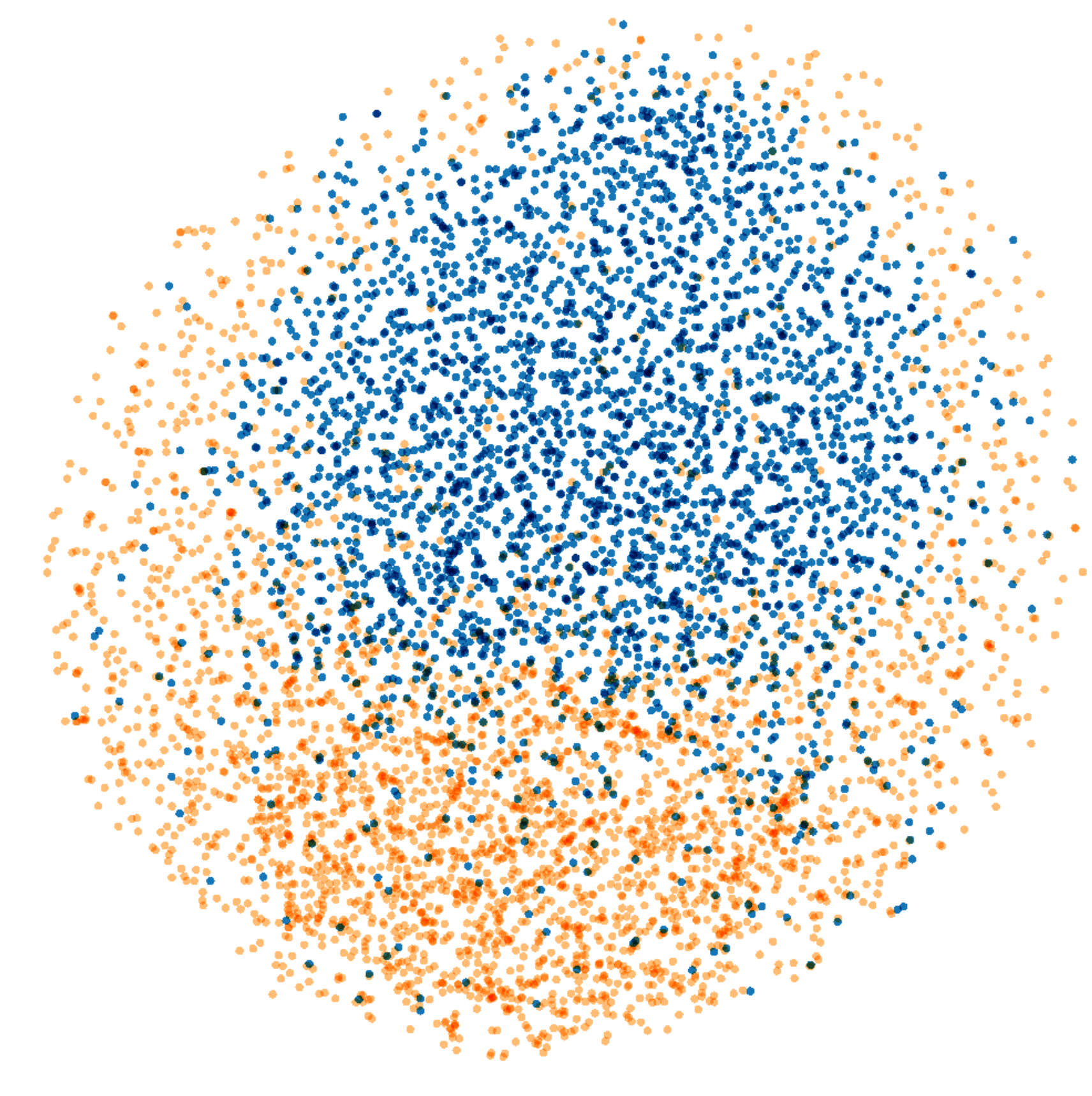}
	\includegraphics[width=.235\textwidth]{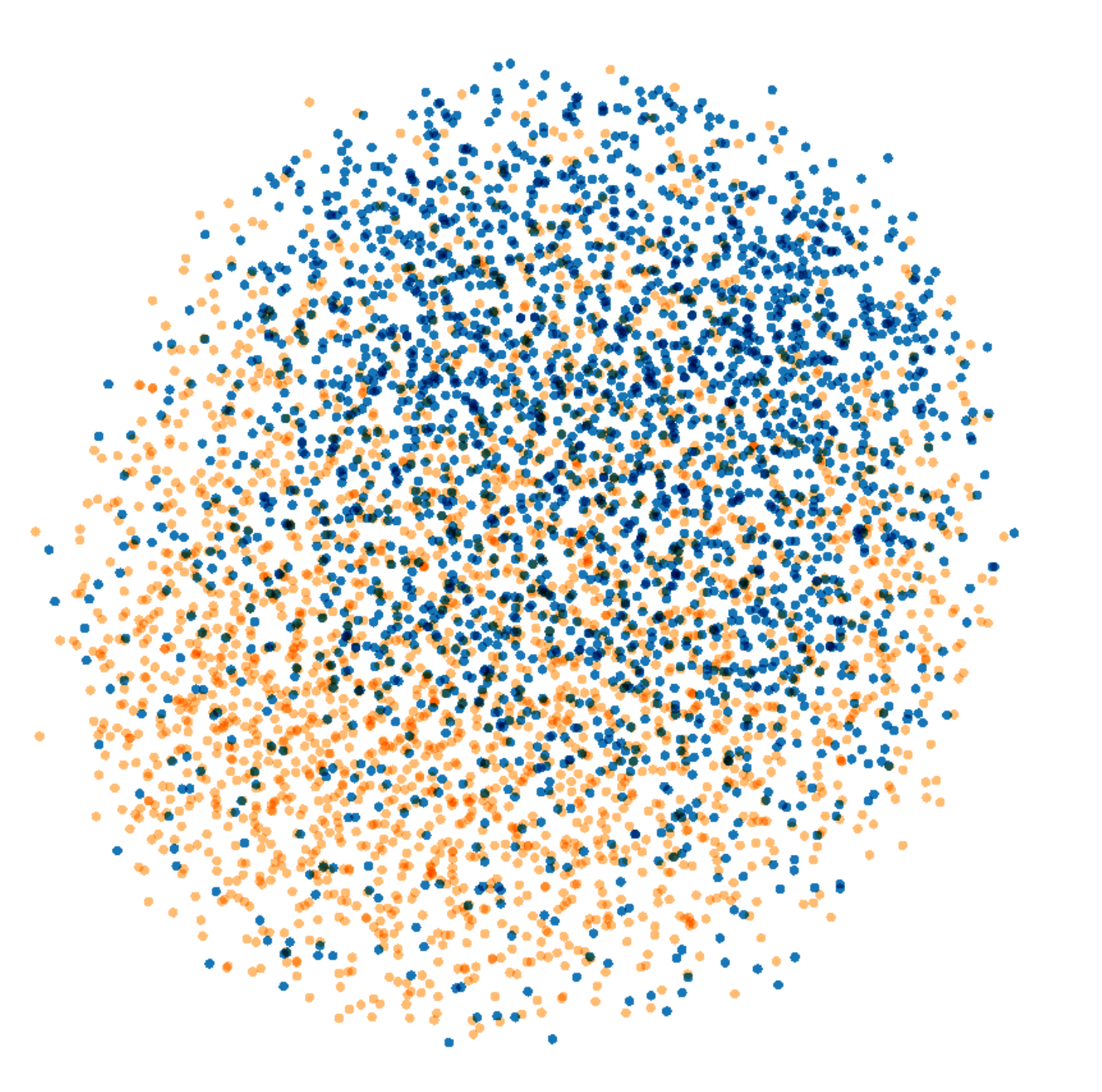}
	\caption{$t$-SNE embeddings of latent codes (\textbf{left}: DeConv-LVM, \textbf{right}: LSTM-LVM) for \emph{BookCorpus} and \emph{arXiv} sentences, which are colored as orange and blue, respectively.}
	\label{fig:embedding}
\end{figure}
	
After the models converge, we randomly sample 5000 sentences from the test set and map their 500-dimensional latent embeddings, $z$, to a 2D vector using $t$-SNE \cite{maaten2008visualizing}.
The embedding plots for DeConv-LVM (left) and LSTM-LVM (right) are shown in Figure~\ref{fig:embedding}. 
For both cases, the plot shape of sampled latent embeddings is very close to a circle, which means the posterior distribution $p(z|x)$ matches the Gaussian prior $p(z)$ well.
More importantly, when we use deconvolutional networks as the decoder, disentangled latent codes for the two writing styles can be clearly observed in the majority of prior space.
This indicates that the semantic meanings of a sentence are encoded into the latent variable $z$, even when we train the model in an unsupervised manner.
On the contrary, the latent codes of LSTM-LVM inferred for different writing styles tend to mix with each other, and cannot be separated as easily as in the case of Deconv-LVM, suggesting that less information may be encoded into the embeddings.
	
\begin{table}
	\centering
	\begin{tabular}{cccccc}
		\hline
		Model &  \# params & Time & KL & Acc    \\
		\hline
		LSTM-LVM          & $\sim$ 16 million & 39m 41s & 4.6 & 91.7 \\
		DeConv-LVM         & $\sim$ 12 million & \bf{8m 23s} & 31.7 & \bf{96.2} \\ 
		\hline
	\end{tabular}
	\caption{Quantitative comparison between latent-variable models with LSTM and deconvolutional networks as the sentence decoder.}
	\label{tab:compare}
\end{table}
	
To better understand the advantages of deconvolutional networks as the decoder in the latent-variable models, we perform a quantitative comparison between the latent codes in DeConv-LVM and LSTM-LVM.
In Table~\ref{tab:compare} we show the number of parameters, training time for 10,000 iterations, and the percentage of KL loss in the total loss for both models.
Moreover, we extract sentence features from each model, and train a linear classifier on top, to distinguish between scientific and informal writing styles.
The sentence embeddings are fixed during training, in order to elucidate the quality of latent codes learned in an unsupervised manner.
1000 sentences are sampled from the training set to learn the classifier and the classification accuracy is calculated on the whole test set.
DeConv-LVM (96.2\%) performs better than LSTM-LVM (91.7\%), again indicating that the the latent codes of DeConv-LVM are more informative.
This observation corresponds well with the fact that the percentage of KL loss in DeConv-LVM (31.7\%) is much larger than in LSTM-LVM (4.6\%), where larger KL divergence loss can be considered as a sign that more useful information has been encoded in the latent variable $z$ \cite{bowman2016generating,yang2017improved}.
Further, we observe that DeConv-LVM has relatively few parameters compared to LSTM-LVM, making it a promising latent-variable model for text.

\subsection{Recognizing Textual Entailment (RTE)}
Motivated by the superior performance of our deconvolutional latent-variable model on unsupervised learning, we further apply it to text sequence matching, in a semi-supervised scenario.
We consider the task of recognizing text entailment on the Stanford Natural Language Inference (SNLI) dataset \cite{bowman2015large}.
%[fixed unsupervised results] \par

\begin{figure}
	\centering
	\includegraphics[width=.4\textwidth]{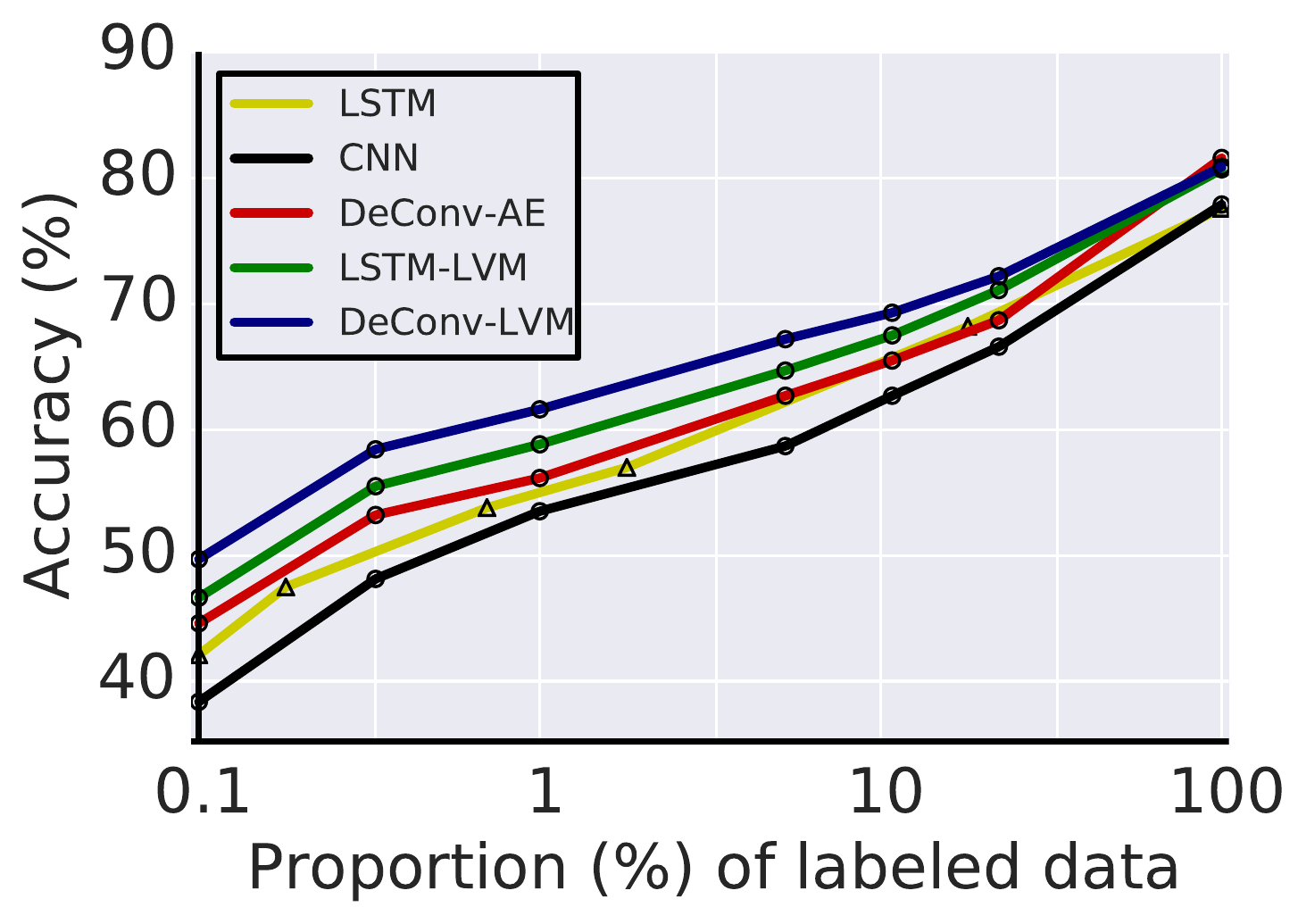}
	\caption{The performance of various models on SNLI dataset, with different amount of labeled data.}
	\label{fig:snli_curve}
\end{figure}

To check the generalization ability of our latent variable learned, we experimented with different amounts of labeled training data (other sentence pairs in the training set are used as unlabeled data).
The results are shown in Figure~\ref{fig:snli_curve}.
Compared to the LSTM baseline models in \cite{bowman2015large} and our basic CNN implementation, both our autoencoder and latent-variable models make use of the unlabeled data and achieve better results than simply train an encoder network, \emph{i.e.}, LSTM, CNN, only with the labeled data.
More importantly, the DeConv-LVM we propose outperforms LSTM-LVM in all cases, consistent with previous observations that the latent variable $z$ in our DeConv-LVM tends to be more informative.
Note that when using all labeled data when training, DeConv-AE (81.6\%) performs a bit better than DeConv-LVM (80.9\%), which is not surprising since DeConv-LVM introduces a further constraint on the latent features learned (close to prior distribution) and may not be optimal when a lot of labeled data are available for training.

\begin{table}[t!]
	\centering
	%\vspace{-2mm}
	%\vskip 0.1in
	%\begin{scriptsize}
	%\bgroup
	%\def\arraystretch{1.5}
	\begin{tabular}{cccccc}
		\hline
		%\hline
		Model & 28k & 59k & 120k   \\
		\hline
		%\hline
		LSTM (\cite{kim2017adversarially})						    &  57.9 & 62.5  & 65.9  \\
		LSTM-AE	(\cite{kim2017adversarially})	             & 59.9 & 64.6 & 68.5  \\
		LSTM-ADAE (\cite{kim2017adversarially})		& 62.5 & 66.8 & 70.9 \\
		\hline
		CNN (random)        & 58.7  & 62.7 & 65.6  \\
		CNN (Glove)         & 60.3 & 64.1 & 66.8  \\
		DeConv-AE         & 62.1 & 65.5 & 68.7  \\
		LSTM-LVM          & 64.7 & 67.5 & 71.1 \\
		DeConv-LVM         & {\bf 67.2} & {\bf 69.3} & {\bf 72.2} \\
		%\hline
        \hline
	\end{tabular}
	%\egroup
		%\vspace{-2mm}
		\caption{Semi-supervised recognizing textual entailment accuracy on SNLI dataset, in percentage. For direct comparison with \cite{kim2017adversarially}. The number of labeled examples is set as 28k, 59k or 120k.}
		\label{tab:snli}
	%\end{scriptsize}
	%\vspace{-4mm}
\end{table}

To directly compare with \cite{kim2017adversarially} on semi-supervised learning experiments, we follow their experiment setup where 28k, 59k, 120k labeled examples are used for training.
According to Table~\ref{tab:snli}, it turns out that our DeConv-AE model is a competitive baseline, and outperform their LSTM-AE results.
Moreover, our DeConv-LVM achieves even better results than DeConv-AE and LSTM-LVM, suggesting that the deconvolution-based latent-variable model we propose makes effective use of unsupervised information.
Further, we see that the gap tends to be larger when the number of labeled data is smaller, further demonstrating that DeConv-LVM is a promising strategy to extract useful information from unlabeled data.

\subsection{Paraphrase Identification}
%[fixed unsupervised results] \par
We investigate our deconvolutional latent-variable model on the paraphrase identification task with the Quora Question Pairs dataset, following the same dataset split as \cite{Wang:2017td}.
We consider cases where 1k, 5k, 10k, 25k labeled examples are used for training.
As illustrated in Table~\ref{tab:quora}, a CNN encoder with Glove pre-trained word embeddings consistently outperforms that with randomly initialized word embeddings, while the autoencoder model achieves better results than only training a CNN encoder, corresponding with findings in \cite{dai2015semi}. \par

More importantly, our latent-variable models show even higher accuracy than autoencoder models, demonstrating that they effectively utilize the information of unlabeled data and that they represent an effective strategy for paraphrase identification task. 
Our DeConv-LVM again performs better than LSTM-LVM in all cases, indicating that the deconvolutional decoder can leverage more benefits from the latent-variable model.
However, we can also see the trend that with larger number of labeled data, the gaps between these models are smaller. 
This may be attributed to the fact that when lots of labeled data are available, discriminative information tends be the dominant factor for better performance, while the information from unlabeled data becomes less important. \par

\begin{table}[t!]
	\centering
	%\begin{scriptsize}
	%\bgroup
	%\def\arraystretch{1.3}
	\begin{tabular}{ccccc}
		\hline
		%\hline
		Model & 1k & 5k & 10k & 25k \\
		\hline
		%\hline
		CNN (random)        & 56.3  & 59.2 & 63.8 & 68.9 \\
		CNN (Glove)         & 58.5 & 62.4 & 66.1 &  70.2 \\
		LSTM-AE     & 59.3 & 63.8 & 67.2 & 70.9 \\
		DeConv-AE         & 60.2 & 65.1 & 67.7 & 71.6 \\
		\hline
		LSTM-LVM          & 62.9 & 67.6 & 69.0 & 72.4 \\
		DeConv-LVM          & {\bf 65.1} & {\bf 69.4} & {\bf 70.5} &  {\bf 73.7} \\
		%\hline
        \hline
	\end{tabular}
	%\egroup
	\caption{Paraphrase identification accuracy on Quora Question Pairs dataset, in percentages.}
	\label{tab:quora}
	%\end{scriptsize}
\end{table}

\section{Related Work}
The proposed framework is closely related to recent research on incorporating NVI into text modeling \cite{bowman2016generating,miao2016neural,xu2017variational,zhang2016variational,serban2017hierarchical}.
\cite{bowman2016generating} presented the first attempt to utilize NVI for language modeling, but their results using an LSTM decoder were largely negative.
\cite{miao2016neural} applied the NVI framework to an unsupervised bags-of-words model.
However, from the perspective of text representation learning, their model ignores word-order information, which may be suboptimal for downstream supervised tasks.
\cite{xu2017variational} employed a variational autoencoder with the LSTM-LSTM architecture for semi-supervised sentence classification.
However, as illustrated in our experiments, as well as in \cite{yang2017improved}, the LSTM decoder is not the most effective choice for learning informative and discriminative sentence embeddings. \par

The NVI framework has also been employed for text-generation problems, such as machine translation \cite{zhang2016variational} and dialogue generation \cite{serban2017hierarchical}, with the motivation to improve the diversity and controllability of generated sentences.
Our work is distinguished from this prior research in two principal respects: (\emph{\romannumeral1})
%\textcolor{blue}{we leveraged NVI framework to text sequence matching tasks, which require semantic reasoning over sentence pairs, for which we proposed to optimize both the generative and discriminative objectives in a joint manner - I DON'T UNDERSTAND WHAT IS INTENDED HERE} \ds{I rephrase it as below:} 
We leveraged the NVI framework for latent variable models to text sequence matching tasks, due to its ability to take advantage of unlabeled data and learn robust sentence embeddings; (\emph{\romannumeral2}) we employed deconvolutional networks, instead of the LSTM, as the decoder (generative) network.
We demonstrated the effectiveness of our framework in both unsupervised and supervised (including semi-supervised) learning cases. \par 

%\smallskip 
\section{Conclusion}
We have presented a latent variable model for matching natural language sentences, with deconvolutional networks as the sequence encoder.
We show that by jointly optimizing the variational lower bound and matching loss, the model is effective at inferring robust sentence representations for determining their semantic relationship, even with limited amount of labeled data. 
State-of-the-art experimental results on two semi-supervised sequence matching tasks are achieved, demonstrating the advantages of our approach.
This work provides a promising strategy towards training effective and fast latent-variable models for text data. \par 

\subsubsection{Acknowledgements} This research was supported in part by ARO, DARPA, DOE, NGA and ONR.

%\bigskip
\small 
\bibliographystyle{aaai.bst} 
\bibliography{vae}
 
\end{document}